\documentclass[lettersize,journal]{IEEEtran}
\usepackage{amsmath,amsfonts}
\usepackage{algorithmicx}
\usepackage{algorithm}
\usepackage{array}
\usepackage[caption=false,font=normalsize,labelfont=sf,textfont=sf]{subfig}
\usepackage{textcomp}
\usepackage{stfloats}
\usepackage{url}
\usepackage{verbatim}
\usepackage{graphicx}
\usepackage{cite}
\usepackage{multirow}
\usepackage{threeparttable}
\usepackage{algpseudocode}

\begin{document}

\title{Imperceptible Physical Attack against Face Recognition Systems via LED Illumination Modulation}

\author{Junbin Fang, Canjian Jiang, You Jiang, Puxi Lin, Zhaojie Chen, Yujing Sun, Siu-Ming Yiu, Zoe L. Jiang
\thanks{Zoe L. Jiang is the corresponding author.}
\thanks{Junbin Fang, Canjian Jiang, You Jiang and Puxi Lin are with the Guangdong Provincial Key Laboratory of Optical Fiber Sensing and Communications, Jinan University, Guangzhou, 510632, China, and also with the Guangdong Provincial Engineering Technology Research Center on Visible Light Communication, and Guangzhou Municipal Key Laboratory of Engineering Technology on VisibleLight Communication, Jinan University, Guangzhou, 510632, China, and also with the Department of Optoelectronic Engineering, Jinan University, Guangzhou, 510632, China(e-mail: tjunbinfang@jnu.edu.cn; canjianjiang@foxmail.com; henanjiangyou@163.com; linpuxi@foxmail.com).} 
\thanks{Zhaojie Chen is with the State Key Laboratory of Modern Optical Instrumentation, Zhejiang University, Hangzhou, 310058, China (e-mail: chenzhaojie1997@foxmail.com).}
\thanks{Yujing Sun and Siu-Ming Yiu are with the Department of Computer Science, The University of Hong Kong, Hong Kong, 999077, China (e-mail:yjsun@cs.hku.hk; smyiu@cs.hku.hk).}
\thanks{Zoe L. Jiang is with the School of Computer Science and Technology, Harbin Institute of Technology, Shenzhen, Shenzhen, 518055, China, and also with the Guangdong Provincial Key Laboratory of Novel Security Intelligence Technologies, Guangdong, 510632, China, and also with the Peng Cheng Laboratory, Shenzhen, 518055, China (e-mail: zoeljiang@hit.edu.cn).}}

\markboth{Journal of \LaTeX\ Class Files,~Vol.~14, No.~8, August~2021}%
{Shell \MakeLowercase{\textit{et al.}}: A Sample Article Using IEEEtran.cls for IEEE Journals}

\IEEEpubid{0000--0000/00\$00.00~\copyright~2021 IEEE}

\maketitle

\begin{abstract}
Although face recognition starts to play an important role in our daily life, we need to pay attention that data-driven face recognition vision systems are vulnerable to adversarial attacks. However, the current two categories of adversarial attacks, namely digital attacks and physical attacks both have drawbacks, with the former ones impractical and the latter one conspicuous, high-computational and inexecutable. To address the issues, we propose a practical, executable, inconspicuous and low computational adversarial attack based on LED illumination modulation. To fool the systems, the proposed attack generates imperceptible luminance changes to human eyes through fast intensity modulation of scene LED illumination and uses the rolling shutter effect of CMOS image sensors in face recognition systems to implant luminance information perturbation to the captured face images. In summary, we present a denial-of-service (DoS) attack for face detection and a dodging attack for face verification. We also evaluate their effectiveness against well-known face detection models, Dlib, MTCNN and RetinaFace , and face verification models, Dlib, FaceNet,  and ArcFace . The extensive experiments show that the success rates of DoS attacks against face detection models reach 97.67$\%$, 100$\%$, and 100$\%$, respectively, and the success rates of dodging attacks against all face verification models reach 100$\%$. 
\end{abstract}


\section{Introduction}

\IEEEPARstart{W}{ith} the rapid development of deep learning technology, face recognition is being widely used in our daily life and under many scenarios, including identity recognition, access control security, financial security, etc\cite{ref1,ref2,ref3}. Meanwhile, the performance of face recognition systems is being continuously enhanced in terms of recognition accuracy as well as recognition speed \cite{ref4,ref5,ref6}. Unsurprisingly, the performance of certain system, GaussianFace \cite{ref7},  have surpassed that of humans. Moreover, the applications of face recognition technology have been further broadened to social and security domains such as trajectory analysis, social media photo tagging, and suspicious person identification in Automated Border Control (ABC) systems \cite{ref8,ref9,ref10,ref34}.
\IEEEpubidadjcol

However, data-driven approaches endogenously suffer from problems such as interpretability and algorithmic black boxes, and do not really obtain causal relationships between samples or features that reflect the nature of the samples, leading to security vulnerabilities in learning-based face recognition systems. Attackers can easily exploit such vulnerabilities and launch adversarial attacks. Note that adversarial attacks can mislead classifiers to produce incorrect predictions by applying small perturbations to the original natural inputs \cite{ref11,ref12}, which are imperceptible to humans, but can cause the current optimal classifier to make incorrect judgments with higher confidence, greatly reducing the success rate of face detection and recognition. For example, attackers can use adversarial attacks to mislead the face recognition system without being detected in order to achieve false authentication and illegal invasion, which not only brings threats to personal information security, property security, and even personal safety, but also poses serious potential risks to urban security and national security  \cite{ref13,ref14}.

Generally, adversarial attacks against face recognition systems can be divided into two main categories: digital adversarial attacks \cite{ref15,ref16} and physical adversarial attacks \cite{ref17,ref18,ref19,ref20}. Digital adversarial attacks assume that the attacker can perform direct pixel-level manipulation of the image input to a learning-based face recognition model, forcing the model to make high-confidence false predictions by directly implanting tiny perturbation patterns. Numerous digital adversarial attack schemes, such as A\textsuperscript{3}GN \cite{ref15}, FLM \cite{ref16}, etc, have been proposed and proved to have very good attack performance, achieving more than 99$\%$ success rate on state-of-the-art face recognition models (e.g., FaceNet \cite{ref21}, ArcFace \cite{ref22}). but digital adversarial attacks require access and privileges to directly write or modify the input image of the face recognition system, which is hard to achieve in the real world, so that the executability and real threat are low. Unlike digital adversarial attacks, physical adversarial attacks physically affect the input image of a face recognition system from the physical world, thereby implanting small perturbations to generate an adversarial example. Compared with digital adversarial attacks, physical adversarial attacks do not directly perturb the original image, so the attack performance does not appear to be as good as digital adversarial attacks, but physical adversarial attacks can be implemented in the real world and have significant or even fatal effects on real systems, so they are receiving more and more attention and researches in recent years.

Currently, physical adversarial attacks on face recognition systems are mainly performed by wearing physical accessories, projections, stickers and other physical methods \cite{ref17,ref18,ref19,ref20} to implant physical perturbations between the face to be detected and the system camera, and form an adversarial image to present to the system camera. Then, the adversarial examples generated during the imaging process of the face recognition system are inputted into a deep neural network (DNN) to achieve the attack purpose, such as AGNs \cite{ref18}, IMA \cite{ref19}, FaceAdv \cite{ref20}, etc. These attack schemes can achieve high attack success rates in physical scenarios under different environmental conditions and generate some threats to real systems, but there are some limitations: (1) low attack inconspicuousness. These types of attacks usually require wearing peripherals or pasting stickers, which are easily detected or perceived by the human eye; (2) low attack executability. These attacks require direct contact with the target object in order to deploy and adjust the attack, which can be easily detected and defended in the real world; (3) low attack generalization and high computational cost. For different attack objects, these attack schemes must compute dedicated perturbation patterns and construct corresponding adversarial examples, which is computationally expensive and poorly adaptable. 

To address these problems, this paper proposes a LED illumination modulation based physical adversarial attack, namingly, LIM, against face recognition systems. First of all, the LED illumination is modulated by high-speed On-Off Keying (OOK) to produce fast flicker beyond the human eye perception frequency. Then,  through the rolling shutter effect of the CMOS image sensor imaging mechanism, the perturbation information carried in the fast illumination flicker can be implanted into the face recognition system imaging process. In the imaging process of face recognition system the perturbation information carried in the fast illumination flicker is implanted into the face image acquired by the system, making the system fails to detect the face and mismatch the face. The proposed approach is conspicuous, highly executable and low computational. Note that the flicking frequency of LED lighting can be changed by adjusting the OOK modulation rate. Additionally, two main types of attacks on the face recognition system are achieved with the proposed LIM by adopting different perturbation patterns, denial of service (DoS) attacks for the face detection stage and dodging attacks for the face feature matching stage. To test the performance of the proposed approach, in a real physical environment, we conduct DoS and dodging attack against three typical face detection models, Dlib \cite{ref23}, MTCNN \cite{ref24}, and RetinaFace \cite{ref25}, and against three advanced face verification models, Dlib \cite{ref23}, FaceNet \cite{ref21}, and ArcFace \cite{ref22}, respectively. The experimental results show that the success rates of DoS attacks reach 97.67$\%$, 100$\%$, and 100$\%$ for the above face detection models , respectively, and those of dodging attacks reach 100$\%$ for all the three face verification models. To sum up, the main contributions are as follows:
\begin{itemize}
\item[$\bullet$] For learning-based face recognition systems, this paper proposes a physical adversarial attack method LIM by modulating LED illumination light, which has the advantages of high inconspicuousness, high executability, and low computational cost.  This work presents a new idea of constructing physical adversarial examples by exploiting the differences between human and computer vision.
\item[$\bullet$] Based on the physical adversarial attack method LIM, this paper further proposes two types of attacks for face recognition systems: denial-of-service attacks and evasion attacks, and also designs adversarial perturbations that can implement attacks for different faces. We formalize the perturbation generation process as an optimization problem, and use a greedy search approach  to solve for the optimal attack parameters (specifically, implantation of perturbation patterns and luminance change control) during the iterative process.
\item[$\bullet$] We evaluate the performance of the proposed physical adversarial attack method in a real physical environment against state-of-art face detection models and face verification models and achieve state-of-art performance.
\end{itemize}

We note that a shorter conference version of this paper appeared in Information Security and Cryptology (2020) \cite{ref31}. Our initial conference paper did not analyze which step of the workflow of the face recognition system the proposed adversarial attack is aimed at. This manuscript adds this analysis, and provides a mathematical generation model and additional experiment for proposed attack.

\section{Related Work}
\begin{table*}[hbtp]
	\begin{center}
		\caption{Summary of typical adversarial example attacks against the face recognition system}
		\label{tab1}
		\renewcommand{\arraystretch}{1.3}
		\begin{threeparttable}
		\begin{tabular}{|c|p{1.5cm}|p{5cm}|c|c|c|p{2cm}|}
			\hline
			Domain & Attack method & Method description & Inconspicuousness & Generalization & Executability & Performance (dataset, model, success rate) \\
			\hline
			\multirow{3}{*}{Digital}  & FLM \cite{ref16} & Manipulating landmark locations to generate adversarial examples & High & Low & * & LFW, FaceNet, \textgreater{}99\% \\
			\cline{2-7}
			~& Dong et al. \cite{ref26} & Presenting an evolutionary attack algorithm to generate adversarial examples in the   decision-based black-box setting & High  & Low & * & LFW, ArcFace, Not given \\
			\cline{2-7}
			~& A\textsuperscript{3}GN \cite{ref15} & Presenting a new GAN with a geometry-based method to generate adversarial examples & High & Low & * & LFW, ArcFace, \textgreater{}99\% \\
			\hline
			\multirow{5}{*}{Physical} & AGNs \cite{ref18} & Generating adversarial accessories in the form of eyeglass frames & Low & Low & Medium & Real faces, Face++, 100\% \\
			\cline{2-7}
			~& FaceAdv \cite{ref20} & Crafting stickers with different shapes attached to the human face & Low & Low & Medium & Real faces, FaceNet, 100\% \\
			\cline{2-7}
			~& IMA \cite{ref19} & Projecting infrared perturbations to human faces & Medium & Low & Medium & Real faces, FaceNet, \textgreater{}77\% \\
			\cline{2-7}
			~& VLA \cite{ref27} & Crafting visible light-based adversarial perturbations and projecting on human faces & High & Low & High & Real faces, FaceNet, \textgreater{}85\% \\
			\cline{2-7}
			~& LIM & Modulating LED illumination to implant perturbations & High & High & High & Real faces,  FaceNet, 100\% \\
			\hline
			\end{tabular}
			\begin{tablenotes}		
				\footnotesize
				\item[*] Only working in the digital world
			\end{tablenotes}
			\end{threeparttable}
		\end{center}
	\end{table*}

Existing adversarial attacks against face recognition systems mainly fall into two  categories: digital adversarial attacks, in which pixel-level manipulation of digital face images is performed to implant imperceptible perturbations to mislead the face recognition system, and physical adversarial attacks, in which small perturbations are implanted in the input image during the imaging stage by physical means to mislead the face recognition system. A summarization of existing adversarial attacks against face recognition systems is shown in Table \ref{tab1}.

{\bf{Digital adversarial attacks:}} These approaches rely on the assumption that the attacker can directly manipulate the input image of the face recognition system and focus more on the effectiveness and imperceptibility of the perturbation and the efficiency of the perturbation computation. In 2018, Dabouei et al. \cite{ref16} proposed a fast landmark manipulation method to generate an adversarial example face image by spatially transforming the original image, which is about 200 times faster than previous geometric attacks and achieves more than 99$\%$ success rate against FaceNet on CASIA-WebFace dataset. However, this method is a white-box attack, which requires the attacker to have full knowledge of the learning-based model parameters and their predictions. For the attack scenarios where the knowledge of face recognition model parameters is completely agnostic, Dong et al. \cite{ref26} proposed an evolutionary attack algorithm to generate decision-based adversarial examples in a black-box environment in 2019. This method performs local geometric modeling of the search direction, while reducing the dimensionality of the search space and improving the search efficiency. However, thousands of queries are needed to ensure the effectiveness, which requires a lot of computing resources. In 2021, Yang et al. \cite{ref15} introduced an Attention Adversarial Attack Generation Network (A\textsuperscript{3}GN), focusing on generating imperceptible adversarial examples to perform white-box, black-box, and targeted attacks that can achieve high success rate attacks against multiple face recognition systems. However, in real-world environments, attackers do not directly manipulate the input images of face recognition systems in most cases, which makes these digital adversarial attack methods poorly executable in physical world scenarios.

{\bf{Physical adversarial attacks:}} In realistic scenarios of face recognition system applications, attackers can only achieve interference at the physical layer where the target object interacts with the system, but cannot manipulate the data transmission inside the system. For this reason, researchers have proposed physical adversarial attacks against realistic face targets. 

The earliest physical adversarial attack against face recognition systems can dated back to 2016, when Sharif et al. \cite{ref18} proposed a method to deceive face recognition systems by wearing printed adversarial eyeglass frame stickers. This method enables an attacker to evade face detection or impersonate another person. However, the color of this adversarial eyeglass frame looks abnormal and the adversarial perturbation is very obvious to human observers. To reduce the perturbation region, Shen et al. \cite{ref20} proposed FaceAdv, where they designed an adversarial sticker production architecture consisting of a sticker generator and a converter to attack face recognition systems by pasting small adversarial stickers in key regions of the face, achieving a high success rate against multiple face recognition systems. However, the reduction of sticker size is still difficult to guarantee the imperceptibility of the adversarial perturbation to human observers. The main idea of this type of physical adversarial attack is to mislead the face recognition system by adding additional accessories or stickers with adversarial perturbation patterns in the face region. Although the human eye cannot detect the presence of adversarial attacks in these patterns, these additional accessories easily attract attention, and this type of attack requires direct contact with the target object, which is not highly executable. In addition, the computational complexity of generating these adversarial perturbation patterns is high, and the adversarial examples obtained by printing also have color or shape distortion, which reduces the robustness of physical adversarial attacks. 

Since the entrance device of a computer vision system (camera) is essentially an optoelectronic sensor, interference from optics can often have critical effects on the computer vision system. In recent years, researchers have also proposed optical physical adversarial attacks that do not require direct contact with the target object. In 2018, Zhou et al. \cite{ref19}  used infrared LEDs to project adversarial perturbation with the form of infrared illumination into a human face to deceive the face recognition system. Although such human eyes-invisible infrared illumination perturbation can be captured by the camera, they be easily filtered out by infrared cut-off filters and will cause serious health problems. To solve the issue, Shen et al. \cite{ref27} proposed the visible light-based attack (VLA), which uses visible light to generate a perturbation frame and a hidden frame that are alternately projected on the human face to deceive the face recognition system. Its imperceptibility relies on the persistence of vision (PoV): if the speed of the frame change between two frames exceeds 25 Hz, the human brain will mix them together and therefore the perturbation cannot be observed. However, it requires specific projection equipment and is difficult to deploy in realistic scenarios. The main idea of this class of physical adversarial attack is to mislead the face recognition system by projecting adversarial illumination perturbations in the face region that can only be captured by the camera. Although the perturbations are enhanced to some extent to be imperceptible in the real world, they require high computational complexity to design specific attack patterns for different faces as well as to compute distortion compensation for the imaging process. 

\section{Details of LIM}
\subsection{Threat Model}
LIM aims to generate physical adversarial examples with both effectiveness and inconspicuousness to deceive state-of-the-art face recognition systems. The characteristics of the proposed LIM are summarized as follows.

{\bf{Black-box attack:}} Since most of the face recognition systems in reality are non-public, it is difficult to obtain their model parameters and data. So  that the black-box attack is more realistic and effective. Our LIM does not need to know the parameters, training data and class labels of the DNN model of the face recognition system, and only implements the attack through the image acquisition process of the target system.

{\bf{Non-targeted attack:}} While implementing LIM, attackers does not need to know in advance which faces exist in the database of face recognition systems, but only needs to mislead the face recognition system to match them as one of the database by reducing the feature distance between face images. Such a non-targeted attack can also deceive the face recognition system and allow the attacker to gain illegal access.

{\bf{Attacker’s capability:}} It is assumed that the attacker has been able to achieve control or placement of the fixture in the physical environment in which it is located at the time the face data is entered into the library during the face image acquisition process.

{\bf{Attacker’s goal:}} The attacker's goal is to make the face recognition system unable to detect and mismatch the input adversarial example face images.

\emph{DoS attack:} The attacker wants to be undetected by the system. In this case, the system will determine "face exists" for face images without added adversarial perturbation, and "face does not exist" for face images with added adversarial perturbation, as follows: 
\begin{equation}
\left\{ {\begin{matrix}
		{f_{1}(X) = 1} \\
		{f_{1}\left( X_{adv} \right) = 0} \\
	\end{matrix}~~} \right.
\end{equation}
where $f_1 \left( \cdot \right)$ is the predicted label output of the face detection model, and ``1" and ``0" correspond to ``face present" and ``face absent", respectively. $X$ is the normal face image, and $X_{adv}$ is the adversarial example face image.

\emph{Dodging attack:} The attacker causes the detected face to be recognized as another person. In this case, the feature matching distance between the face image of the attacker and the legitimate user with added adversarial perturbation will be less than the verification threshold, as follows: 
\begin{equation}
d\left( {f_{2}\left( X_{adv} \right),f_{2}\left( U_{adv} \right)} \right) \leq \delta
\end{equation}
Where $f_2 \left( \cdot \right)$ is the face feature vector output of the face verification model, $d\left( \cdot \right)$ is the face feature distance between two face samples, $X_{adv}$ is the face image of the attacker with adversarial perturbation added, $U_{adv}$ is the face image of the legitimate user with adversarial perturbation added, and $\delta$ is the verification threshold.

\subsection{The Proposed LIM Scheme}
\begin{figure*}[!t]
	\centering
	\includegraphics[width=6.5in]{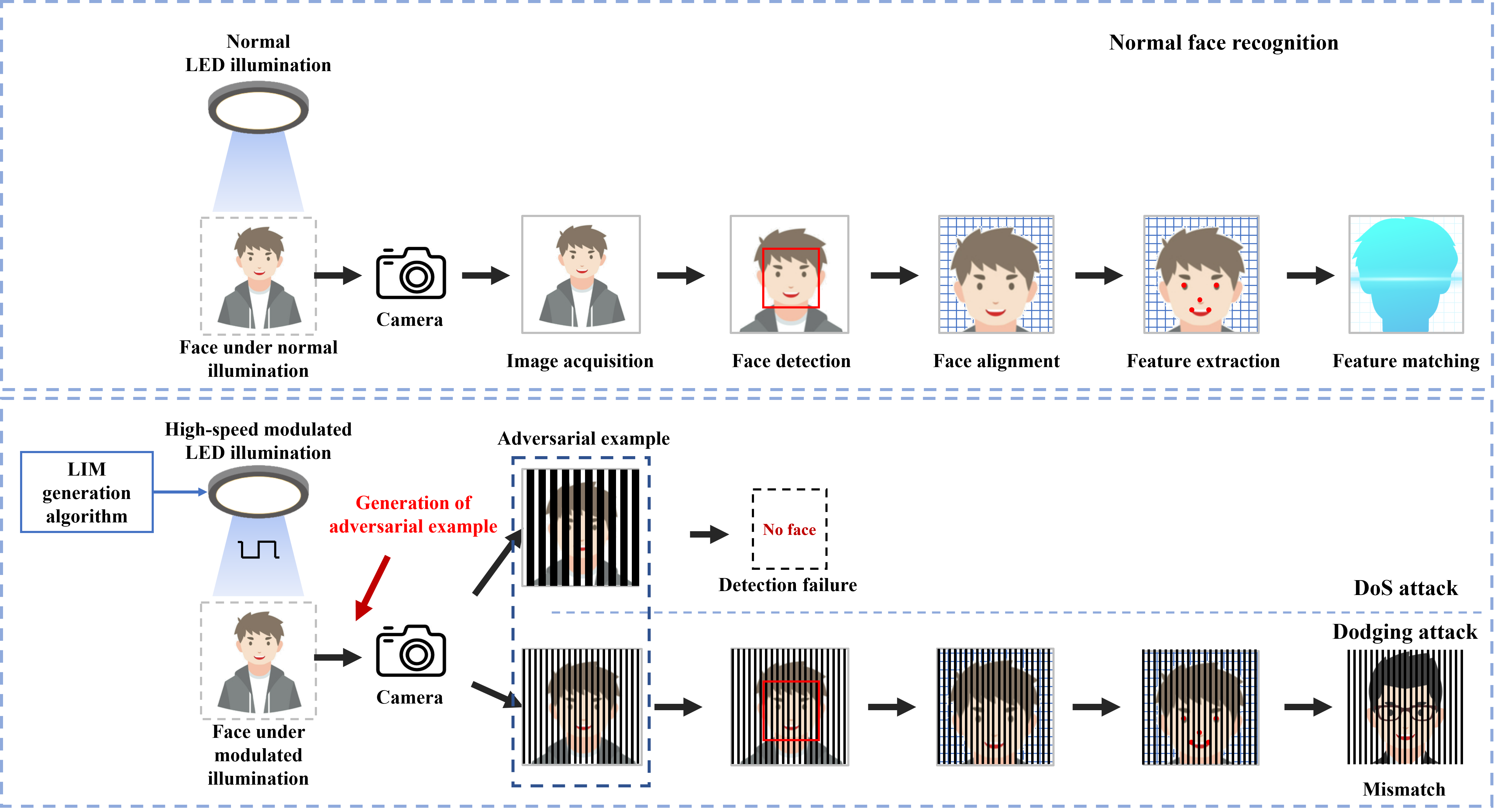}
	\caption{The proposed LIM scheme.}
	\label{fig_1}
\end{figure*}
The workflow of a conventional face recognition system is shown in the upper part of Fig. \ref{fig_1}. In the image acquisition stage, the camera of the face recognition system images a human face under normal illumination conditions to obtain the original face image, and then goes through the stages of face detection, face alignment, feature extraction and feature matching to finally recognize the acquired face image.

The physical adversarial attack scheme proposed in this paper is shown in the lower part of Fig. \ref{fig_1}. First, in the image acquisition stage of the system, we use the adversarial example generation algorithm (see \uppercase\expandafter{\romannumeral3}-\emph{D} for details) to modulate the lighting LED lamps at high speed so that the ambient lighting contains high-frequency light brightness flicker, which are beyond the flicker frequency range that can be perceived by the human eye. So that they are not detected by the human eye. Due to the rolling shutter effect of the CMOS image sensor imaging mechanism, when the camera of the face recognition system images the face under the modulated lighting conditions, these high-frequency light flickers are captured and automatically generate a small adversarial perturbation pattern corresponding to the high-frequency light flickers. And this perturbation is implanted in the captured original face image, thus realizing the generation of adversarial examples and being read in by the face recognition system. According to the different adversarial perturbation patterns designed, the attack scheme proposed in this paper has two different attack modes: DoS attack for the face detection stage and dodging attack for the face feature matching stage. The specific steps of this scheme are as follows.

(1) \emph{Modulating the LED lighting:} In order to improve the inconspicuousness of the attack in physical scenes, the LIM physical adversarial perturbation generation utilizes LED lighting used in everyday lighting without any special light sources or projection devices. LED lighting has the outstanding advantage of being green and energy efficient and has become the current mainstream lighting source. The LED light-emitting chip also has physical characteristics such as fast response, adjustable brightness, and support for high-speed modulation. Since the human eye does not perceive light flicker more than 200Hz \cite{ref30}, when we conduct high-speed OOK modulation (i.e., using binary information ``1" or ``0" to control the light) at a frequency close to 2000Hz for LED lamps, the human eye will not perceive the high-frequency light flicker, but only see the uniform and continuous brightness of the lighting.

The modulation parameters of the LED luminaire can be calculated according to the proposed attack mode. We optimize the design of the perturbation pattern to generate adversarial examples. And the detailed model will be described in section \uppercase\expandafter{\romannumeral3}-\emph{C}.
\begin{figure}[!t]
	\centering
	\includegraphics[width=3in]{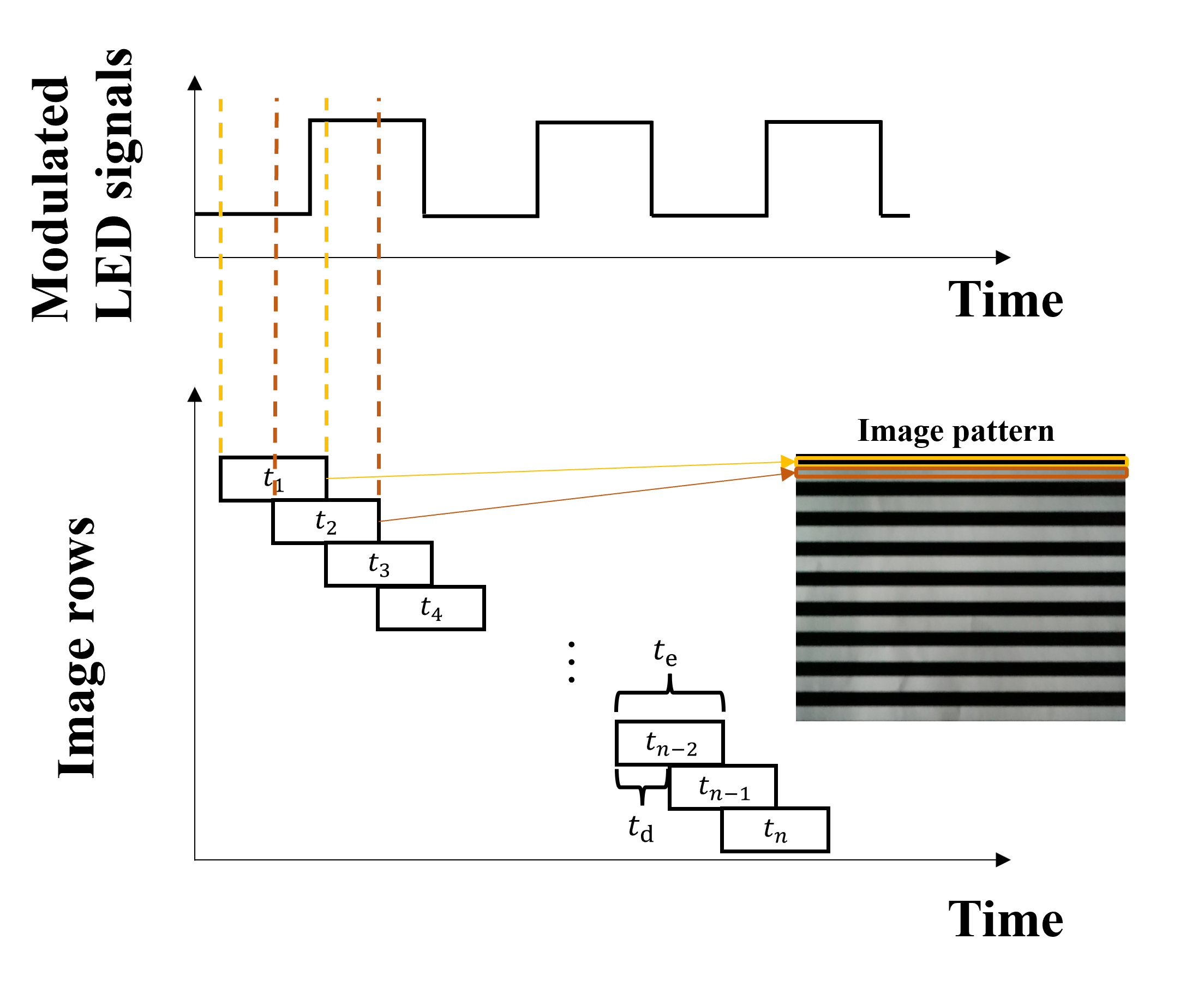}
	\caption{The rolling shutter effect.}
	\label{fig_2}
\end{figure}

(2) \emph{Generating physical adversarial examples:} Most of the cameras of face recognition systems use CMOS image sensors, which adopt the rolling shutter progressive exposure mode on the exposure imaging mechanism, i.e., pixels are exposed line by line with a constant interline delay $t_d$  until all pixels are exposed. The high-frequency light flicker generated by step (1) is not perceived by the human eye but can be captured by the camera of the face recognition system. As shown in Fig. \ref{fig_2}, in time quantum $t_1$, when the CMOS image sensor exposes the first line of pixels, the LED illumination is mostly in the off light state, and the line pixel value read at this time is black. While in time quantum $t_2$, when the CMOS image sensor exposes the second line of pixels, the LED illumination is mostly in the on light state, and the line pixel value read at this time is the normal image value. And so on, the acquired face image will carry the bright and dark fringes corresponding to the high-frequency light flashing. These fringes can be considered as adversarial perturbations superimposed on the original face image. And the face image superimposed with adversarial perturbation fringes becomes the adversarial example and is read by the face recognition system, such as the adversarial example image in the image acquisition step in Fig. \ref{fig_1}. Note that in the case of Fig. \ref{fig_1}, since the camera of the face recognition system usually takes pictures at a vertical angle, the rows of pixels (long edges) in the acquired image are also vertical, so the implanted adversarial perturbations are also presented as vertical fringes.
   
Depending on the designed adversarial perturbation patterns, the attack scheme proposed in this paper can have two different attack strategies: DoS attack for the face detection stage and dodging attack for the feature matching stage.

(3) \emph{DoS attack for face detection stage:} DoS attack is an attack on the face detection process of the face recognition system, which aims to make the face recognition system unable to detect the face, thus leading to the failure of the system and the termination of the subsequent process. The key to achieving this purpose is how to disrupt the model's detection of the five facial features of the face. The intuitive idea is to mask all five facial features of the face, which means that the dark fringes are wide enough to mask the five facial features of the face or even cover the face. However, this approach is more extreme, and wider dark fringes require a longer duration of LED fixture extinction, which will result in LED flicker being perceived by the human eye. Therefore, our DoS attack is designed to interrupt the continuous detection of five facial features by presenting bright and dark fringes of appropriate width in the captured face image with LED illumination modulated at a lower frequency, such as strong interference covering 1/3 or 1/4 of the area in the middle of the five facial features with wider dark fringes. As shown in the DoS attack phase in Fig. \ref{fig_1}, an attacker can implement DoS attack to reject the legitimate user from the face recognition system or evade face detection.

(4) \emph{Dodging attack for feature matching stage:} Dodging attack is an attack on the feature extraction and feature matching process of the face recognition system, aiming to narrow the feature matching distance between different faces to make it lower than the verification threshold of the face recognition system, thus causing the system to recognize the attacker as another person. Since the wider bright and dark fringes interference will cause the face recognition system to fail in face detection and cannot complete the subsequent face alignment and feature extraction. Therefore, the bright and dark fringes interference of the dodging attack should not obscure the five facial features of the face too much, but focus on changing the expression of the feature information, such as adding ``pseudo-contour" to the facial features through narrower dark fringes, resulting in the deviation of the facial landmarks and the change of the corresponding feature vectors. The dodging attack is designed to add a lot of repetitive and useless information to the face features by presenting narrow dark fringes in the captured face image with LED illumination modulated at a higher frequency. As shown in the dodging attack phase of Fig. \ref{fig_1}, the images of two different people are implanted with these useless narrow bright and dark fringes information, and the feature matching distance of their face images will drop below the verification threshold and mislead the face recognition system to match them as the same person.  An attacker can use dodging attack to achieve false authentication or trespassing.

\subsection{Adversarial Attack Model}
\begin{figure*}[!t]
	\centering
	\includegraphics[width=7in]{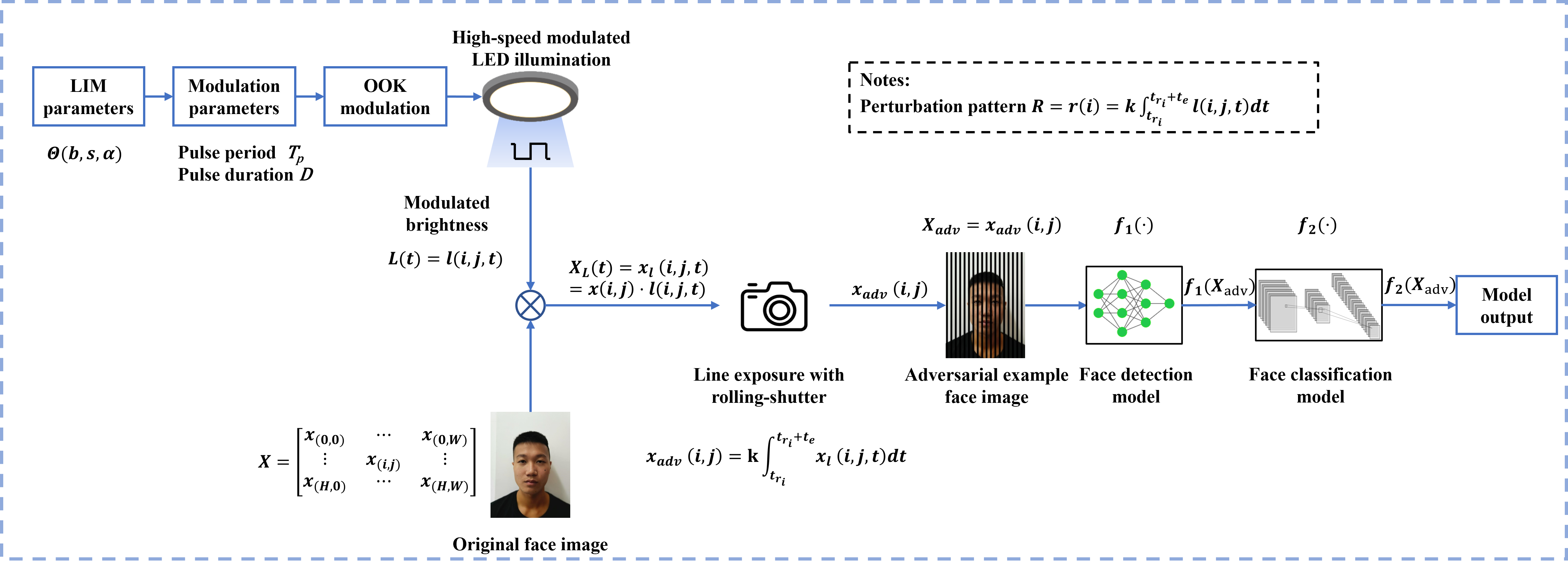}
	\caption{LIM adversarial perturbation pattern generation model.}
	\label{fig_3}
\end{figure*}

The above LIM scheme can be abstracted as a mathematical model as shown in Fig. \ref{fig_3}.

(1) Let the original portrait data $X$ falling into the FOV area of the camera be:
\begin{equation}
X = \begin{bmatrix}
	{x(0,0)} & \cdots & {x\left( {0,W} \right)} \\
	\vdots & {x\left( {i,j} \right)} & \vdots \\
	{x\left( {H,0} \right)} & \cdots & {x\left( {H,W} \right)} \\
\end{bmatrix}
\end{equation}
where $W$ and $H$ are width and height of the image , and $x(i, j)$ is the intensity of the original portrait at pixel point location $(i, j)$.

(2) Let the set of attack parameter set of LIM be $\Theta$, including the width $b$, interval $s$ and tilt angle $\alpha$ of the perturbation fringe in the adversarial perturbation pattern. The width $b$ and interval $s$ are related to the parameters of the OOK modulated pulse electrical signal (including pulse repetition period $T_p$, pulse duty cycle $D$) that control the LED illumination modulation. Their corresponding relationships can be formulated as:
\begin{equation}
b = \frac{T_{p}D}{t_{d}}
\end{equation}
\begin{equation}
s = \frac{T_{p}\left( {1 - D} \right)}{t_{d}}
\end{equation}
where $t_d$ is the interline delay of progressive exposure. 

(3) The modulation pulse parameters are OOK modulated to control the LED lighting to produce high-speed light flicker, thus making the LED illumination carry a perturbation signal. Let the illumination light signal $L(t)$ falling into the FOV region of the camera with time $t$ be:
\begin{equation}
L(t) = \begin{bmatrix}
	{l\left( {0,0,t} \right)} & \cdots & {l\left( {0,W,t} \right)} \\
	\vdots & {l\left( {i,j,t} \right)} & \vdots \\
	{l\left( {H,0,t} \right)} & \cdots & {l\left( {H,W,t} \right)} \\
\end{bmatrix}
\end{equation}
where $l(i,j,t)$ is the illumination brightness value of the illumination light at moment $t$ at pixel location $(i,j)$.

(4) According to the principle of optical imaging, the intensity of the portrait data is linearly related to the illumination, so  that the illumination light signal is multiply superimposed with the portrait data. That is,  the target portrait data $X_L (t)$ after mixed illumination is obtained as the matrix Hadamard product  of the original portrait data $X$ and the illumination light signal $L(t)$.
\begin{equation}
X_{L}(t) = X \circ L(t) = \left\lbrack {x\left( {i,j} \right) \cdot l\left( {i,j,t} \right)} \right\rbrack = \left\lbrack {x_{l}\left( {i,j,t} \right)} \right\rbrack
\end{equation}

(5) In the next step, the mixed portrait data $X_L (t)$ is passed through the camera and combined with the imaging mechanism of the camera to obtain the adversarial example $X_{adv}$. The exposure imaging process of the CMOS image sensor is actually the accumulation and photoelectric conversion of the incident luminous flux. And because the camera uses the row-by-row exposure imaging mechanism of the rolling shutter, the CMOS image sensor will expose the pixels row by row at a certain time interval. The time interval of the exposure is determined by the speed of the shutter and line pixel value reading time, so the exposure start time $t_{r_i}$ for each pixel line of the image varies in turn. Let the exposure duration of each line be $t_e$ and the conversion gain of the camera be $k$. The pixel value $x_{adv}(i,j)$ of each pixel point is obtained as the integral of the mixed portrait data $X_L (t)$ within the exposure duration as follows: 
\begin{equation}
\begin{matrix}
	{x_{adv}\left( {i,j} \right) = k{\int_{t_{r_{i}}}^{t_{r_{i}} + t_{e}}{x_{l}\left( {i,j,t} \right)dt}}} \\
	{~~~~~~~~~~~~~~~~~~~~ = x\left( {i,j} \right) \cdot k{\int_{t_{r_{i}}}^{t_{r_{i}} + t_{e}}{l\left( {i,j,t} \right)dt}}} \\
\end{matrix}
\end{equation}

Therefore, the adversarial example face image $X_{adv}$  is: 
\begin{equation}
X_{adv} = \left\lbrack {x_{adv}\left( {i,j} \right)} \right\rbrack = \left\lbrack {k{\int_{t_{r_{i}}}^{t_{r_{i}} + t_{e}}{x_{l}\left( {i,j,t} \right)dt}}} \right\rbrack
\end{equation}

As shown in Eq. (7), the original portrait data $X$ contained in the hybrid portrait data $X_L (t)$ is independent of time $t$ and can be extracted outside the integral equation. Therefore, the adversarial example face image $X_{adv}$ can be expressed as:
\begin{equation}
X_{adv} = X \circ R = \left\lbrack {x\left( {i,j} \right) \cdot k{\int_{t_{r_{i}}}^{t_{r_{i}} + t_{e}}{l\left( {i,j,t} \right)dt}}} \right\rbrack
\end{equation}
where $R$ is defined as the adversarial perturbation pattern introduced by this attack scheme, generated by the LED illumination modulation in conjunction with the camera exposure mechanism.
\begin{equation}
R = r(i) = k{\int_{t_{r_{i}}}^{t_{r_{i}} + t_{e}}{l\left( {i,j,t} \right)dt}}
\end{equation}

Therefore, in the case of constant camera exposure parameters, by adjusting the modulation parameters, it is possible to design different adversarial perturbation patterns $R$ to achieve different attack modes. Meanwhile, for different camera exposure parameters, only simple adjustment of the modulation parameters is required to complete the adaptation and improve the universality of the attack scheme.

(6) Let $f_1 (\cdot)$ be the face detection model output (i.e., the face detection output label value) in the face recognition system, the objective of the proposed DoS attack is to make the label value of the adversarial example output through this discriminator converge to a value of $0$, which means ``face region does not exist", i.e., let $f_1 (X_{adv})=0$. The optimization formula is defined as the maximized difference between the predicted label value of the adversarial example $f_1 (X_{adv})$ and the true label value of the original example $y$.
\begin{equation}
L_{1} = {\underset{\Theta{({b,s,\mathit{\alpha}})}}{\arg\max}{~\left\lbrack {{y - f}_{1}\left( X_{adv} \right)} \right\rbrack^{2}}}
\end{equation}

In our tests, the true label value $y=1$. And if the value of $L_1$ is $1$, the DoS attack is considered successful.

(7) Let $f_2 (\cdot)$ be the output of the face verification model (i.e., the face feature matching output vector value) in the face recognition system. The purpose of the proposed dodging attack in this paper is to narrow the face feature distance $L_2$ between two face images so that it is lower than the determination threshold of the face verification model, thus scrambling the face feature matching output label value.
\begin{equation}
L_{2} = {\underset{\Theta{({b,s,\alpha})}}{\arg\min}~\sqrt{\left\lbrack {{f_{2}\left( X_{adv} \right) - f}_{2}\left( U_{adv} \right)} \right\rbrack^{2}}}
\end{equation}

If $L_2<\delta$, the dodging attack is considered successful.

\subsection{Optimal Search Algorithm for Adversarial Perturbation Parameters}
As mentioned above, in the LIM scheme, our optimization problem is actually to find the optimal set of adversarial perturbation pattern parameters $\Theta(b,s,\alpha)$ for the optimization objective of Eq. (12) or (13), which aims to lead to the inability of the face recognition system to detect faces in the captured images or to mismatch faces in the images, for which we propose the LIM optimal search algorithm.

In a practical face recognition application scenario, the attacker has no access to the knowledge of the target model and can only obtain the detection labels of the given input images and the distance of face features between different input images. This search algorithm takes the proposed attacked face image $X$ as input and uses the detection labels and face feature distance results provided by the target model to find effective adversarial perturbation parameters in the defined search space of fringe width $b$, interval $s$ and tilt angle $\alpha$ in a greedy search manner. We use $b_{max}$, $s_{max}$, and $\alpha_{max}$ to set the maximum number of steps of the search process, while the predicted output of the model is used as the variable of the optimization Eq. (12) or (13). We solve the results of these equations and compare them with the threshold value for judgment. If the results meet the judgment conditions, the effective adversarial perturbation parameters will be retained. The search ends when the loop reaches the maximum number of steps and outputs a list of adversarial perturbation parameters containing the fringe width $b$, interval $s$ and tilt angle $\alpha$. The pseudocode of the search algorithm is shown in Algorithm \ref{alg:alg1}. 

\begin{algorithm}[H]
	\caption{Pseudocode of LIM.}\label{alg:alg1}
	\begin{algorithmic}[1]
		\Require{
			Input image $X, U$; Maximum iterative number $c$; Maximum width $b_{max}$; Maximum interval $s_{max}$; Maximum angle $\alpha_{max}$; Detector $f_1$; Feature extractor $f_2$.}
		\Ensure{
			Adversarial perturbation parameter $\theta_1$, $\theta_2$.}
		\For{$i\gets 1$ \textbf{to} $c$}
		\State\textbf{Initialize} $\theta\sim\Theta \left(b, s, \alpha \right), \delta$
		\For{$b\gets 1$ \textbf{to} $b_{max}$, $s\gets 1$ \textbf{to} $s_{max}$}
		\For{$\alpha\gets 0$ \textbf{to} $\alpha_{max}$}
		\State $\theta_1 \gets \theta\left(b, s, \alpha \right)$	
		\State $X_{adv} \gets X_{\theta_1}$	
		\State $L_1 \gets\left\lbrack {1 - f}_{1}\left( X_{adv} \right) \right\rbrack^{2}$
		\If {$L_1 > 0}$
		\State\textbf{return} $\theta_1$
		\EndIf
		\State $\theta_2 \gets \theta\left(b, s, \alpha \right)$	
		\State $X_{adv} \gets X_{\theta_2}, U_{adv} \gets U_{\theta_2}$	
		\State $L_2 \gets \sqrt{\left\lbrack {{f_{2}\left( X_{adv} \right) - f}_{2}\left( U_{adv} \right)} \right\rbrack^{2}}$
		\If {$L_2 > \delta}$
		\State\textbf{return} $\theta_2$
		\EndIf
		\EndFor
		\EndFor
		\EndFor
	\end{algorithmic}
\end{algorithm}

The algorithm finally calculates two sets of effective LIM adversarial perturbation parameter lists $\theta _1=[(b_{11},s_{11},\alpha_{11}),(b_{12},s_{12},\alpha_{12}),…,(b_{1n},s_{1n},\alpha_{1n})]$ and $\theta_2=[(b_{21},s_{21},\alpha_{21} ),(b_{22},s_{22},\alpha_{22}),…,(b_{2n},s_{2n},\alpha_{2n})]$ used for DoS attack and dodging attack respectively.

\section{Experiments and Discussions}
To evaluate the effective of the proposed LIM, we conduct LIM based DoS attack and dodging attack in a real physical environment against state-of-the-art face detection models Dlib \cite{ref23}, MTCNN \cite{ref24}, and RetinaFace \cite{ref25}, and state-of-the-art face verification models, Dlib \cite{ref23}, FaceNet \cite{ref21}, and ArcFace \cite{ref22}, respectively.
 
(1) {\bf{Face detection models}}

\emph{Dlib} \cite{ref23}\emph{:} Detects faces and locates the positions of facial features using the histogram of orientation gradients (HOG) to obtain the 68 feature point positions of faces. We test that the the dataset achieves 99.25$\%$ detection accuracy on a subset of LFW \cite{ref29} dataset. 

\emph{MTCNN} \cite{ref24}\emph{:} Uses P-Net for fast candidate window generation, R-Net for filtering selection of high-precision candidate windows, and O-Net for generating the final bounding box with 5 key feature points of faces to conduct fast and efficient face detection, achieving 85.1$\%$ detection accuracy on the well-known public benchmark face dataset WiderFace \cite{ref28} for face detection.

\emph{RetinaFace} \cite{ref25}\emph{:} Performs pixel-level face localization at various face scales using a joint externally supervised and self-supervised multi-task learning framework, while predicting face scores, bounding boxes, 5 face key feature points and 3D locations, achieving 96.3$\%$ detection accuracy on the well-known public benchmark face dataset WiderFace \cite{ref28}  for face detection.

(2) {\bf{Face verification models}}

\emph{Dlib} \cite{ref23}\emph{:} extracts face feature vectors by using ResNet features, and the similarity of faces is judged by solving the Euclidean distance between the two sets of feature vectors.

\emph{FaceNet} \cite{ref21}\emph{:} directly learns the mapping from face images to compact Euclidean space, where the Euclidean distance directly corresponds to the face similarity metric.

\emph{ArcFace} \cite{ref22}\emph{:} maximizes the verification bounds in angular space and introduces an additional angular edge loss to obtain highly discriminative features.

The accuracy of Dlib \cite{ref23}, FaceNet \cite{ref21}, and ArcFace \cite{ref22} on the well-known public benchmark face dataset LFW  \cite{ref29} for face verification is 99.38$\%$, 99.65$\%$, and 99.65$\%$, respectively. We also collected 120 face images of three volunteers at different shooting distances under normal illumination conditions for validating the actual performance of the models. The performance of three face verification models on LFW  \cite{ref29} dataset and volunteer face dataset is shown in Table \ref{tab2}.

\begin{table}[hbtp]
	\begin{center}
		\caption{The verification accuracy of face verification models on LFW and volunteer dataset}
		\label{tab2}
		\renewcommand{\arraystretch}{1.3}
		\begin{tabular}{|l|p{3cm}|p{3cm}|}
		\hline
		Model   & Verification accuracy on LFW & Verification accuracy on volunteer dataset \\
		\hline
		Dlib    & 99.38\%                        & 100\%                                        \\
		\hline
		FaceNet & 99.65\%                        & 100\%                                        \\
		\hline
		ArcFace & 99.65\%                        & 100\%                                        \\
		\hline                
		\end{tabular}
	\end{center}
\end{table}

\subsection{Experimental Setup}
\begin{figure}[hptb]
	\centering
	\includegraphics[width=3in]{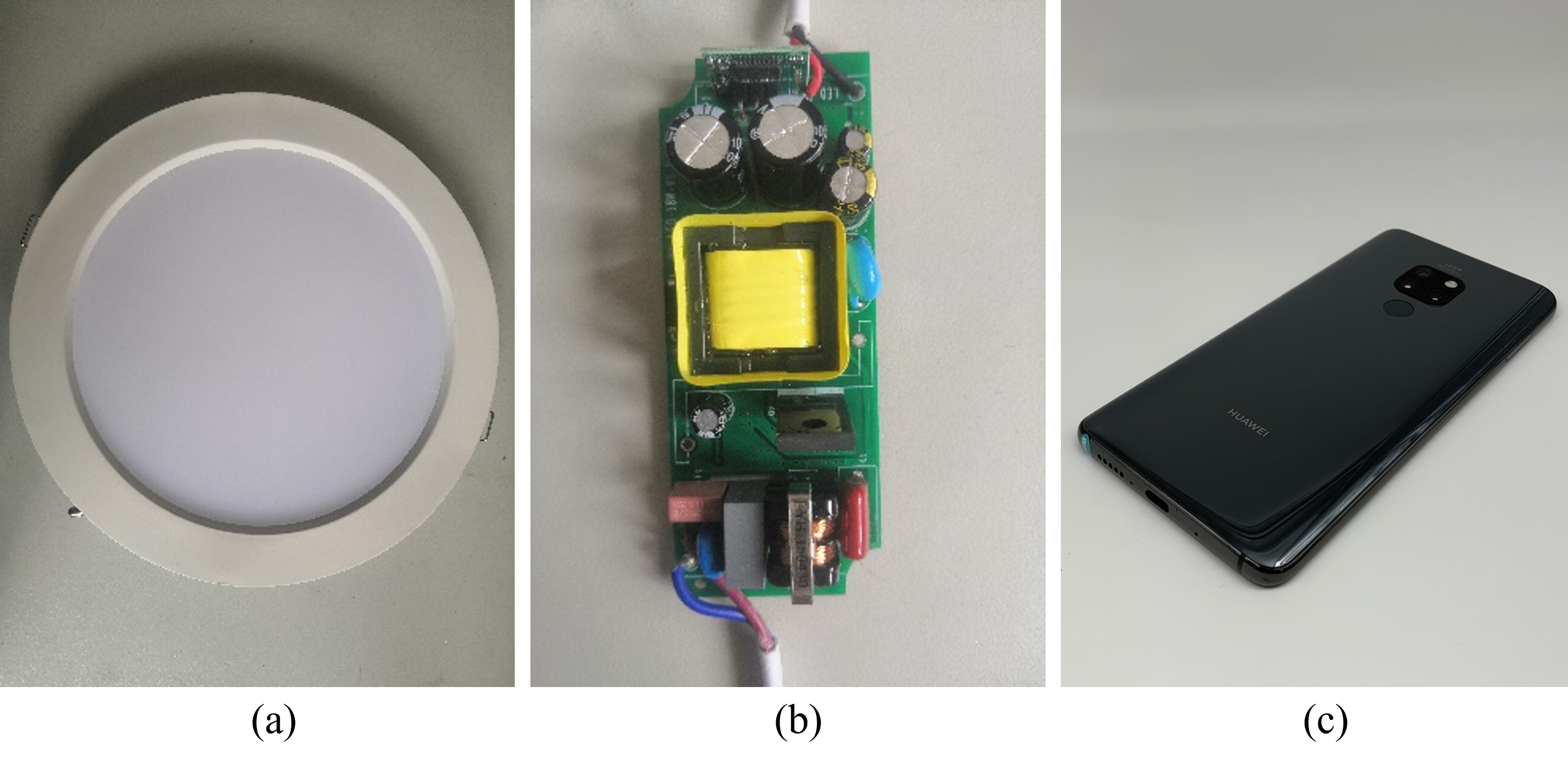}
	\caption{Experimental devices. (a) LED lamp; (b) High-speed modulator; (c) HUAWEI Mate 20 mobile phone.}
	\label{fig_4}
\end{figure}
The devices used are shown in Fig. \ref{fig_4}, including a commercially available LED lamp with a diameter of 17.5cm, a power of 18W as the illumination source, and a LED lamp with a self-developed on-off-keyed high-speed modulator. The modulation pulse repetition period $T_p$ is adjustable from 100$\mu$s to 2000$\mu$s, the pulse duty cycle $D$ is set to 0.5, and accordingly, the width of the modulating pulse can be adjusted from 50$\mu$s to 1000$\mu$s. By adjusting the modulation pulse parameters, the high-frequency flicker of the lighting signal can be controlled to produce different adversarial perturbation patterns. We use a Huawei Mate 20 mobile phone equipped with CMOS image sensor as the camera of the face recognition system. And the test sample of the physical adversarial attack experiment is obtained by capturing the face image under LED illumination. The image resolution is 960 × 1280, the shutter speed parameter is set to 1/4000sec, and the sensitivity (ISO) is set to 3200.

\begin{figure*}[!t]
	\centering
	\includegraphics[width=6in]{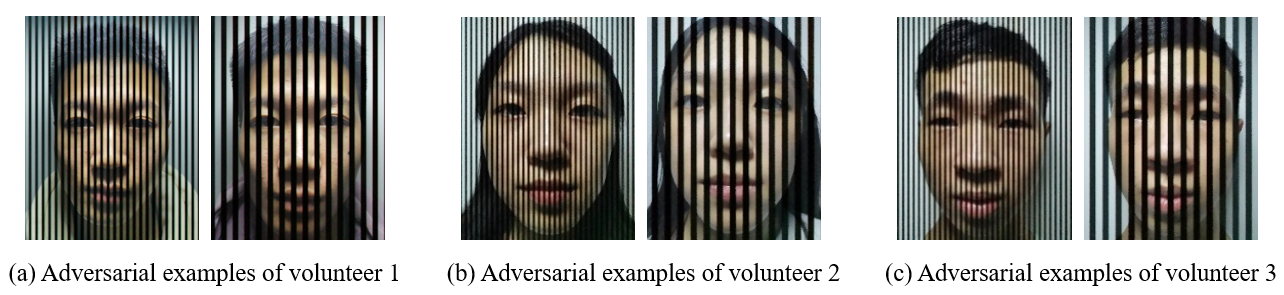}
	\caption{Adversarial examples of LIM.}
	\label{fig_5}
\end{figure*}

In the experiments of this paper, we first set different adversarial perturbation parameters (fringe width $b$, interval $s$ and tilt angle $\alpha$). Then, we convert them into modulated pulse parameters (pulse repetition period $T_p$ and pulse duty cycle $D$) for the LED lamp. Finally, the LIM adversarial example dataset is collected by the camera. Three volunteers (including two males and one female) between the ages of 20 and 22 volunteered to participate in the collection, in which each volunteer collected 698 adversarial examples, and a total of 2094 adversarial examples were collected. Adversarial examples of LIM from three volunteers are shown in Fig. \ref{fig_5}.

\subsection{Experimental Results of DoS Attack}
A successful DoS attack on the face detection stage means that the face recognition system cannot detect the face features and the face detection function is disabled. If the number of samples that can successfully pass face detection before the attack is $n_b$, and the number of adversarial examples that cannot detect faces after DoS attack is $n_a$, we define the success rate of DoS attack as: 
\begin{equation}
r_{DoS} = ~\frac{n_{a}}{n_{b}}
\end{equation}

In total, we measure the success rate of DoS attack under different pulse repetition periods, different shooting distances and different tilt angles.

(1) {\bf{DoS attack with different pulse repetition periods.}}

We collected 100 images each of the faces of three volunteers under normal LED illumination and modulated LED illumination with pulse repetition periods of 1000$\mu$s, 1200$\mu$s, 1400$\mu$s, 1600$\mu$s, 1800$\mu$s, and 2000$\mu$s, for a total of 2100 images in the experimental setting described in \uppercase\expandafter{\romannumeral4}-\emph{A}, with a fixed camera-to-face distance of 18cm.

The performance of DoS attack with modulated LED illumination of different pulse repetition periods is illustarted in Table \ref{tab3}. For each face detection model, we obtain the success rate metrics of DoS attack in the form of a test-by-test comparison of normal samples and adversarial examples according to Eq. 14. To be specific, for face images under normal illumination, the detection accuracy of all three face detection models is 100$\%$.  While the modulated LED illumination with all the different pulse repetition periods tested poses a real threat to all three typical face detection models, and the attack success rates are higher than 92$\%$.
\begin{table}[hbtp]
	\caption{The success rate of DoS attack with different pulse periods}
	\label{tab3}
	\renewcommand{\arraystretch}{1.3}
	\begin{tabular}{|l|p{0.7cm}|p{0.7cm}|p{0.7cm}|p{0.7cm}|p{0.7cm}|p{0.7cm}|}
		\hline
		\multirow{2}{*}{Target model} & \multicolumn{6}{c|}{Pulse period}                        \\
		\cline{2-7}
		~                             & 1000$\mu$s  & 1200$\mu$s  & 1400$\mu$s  & 1600$\mu$s  & 1800$\mu$s & 2000$\mu$s  \\
		\hline
		Dlib                                              & 92.77\% & 97.33\% & 96.67\% & 97.33\% & 96\%   & 97.67\% \\
		\hline
		MTCNN                                             & 100\%   & 100\%   & 100\%   & 100\%   & 100\%  & 100\%   \\
		\hline
		RetinaFace                                        & 100\%   & 100\%   & 100\%   & 100\%   & 100\%  & 100\%  \\
		\hline
	\end{tabular}
\end{table}

(2) {\bf{DoS attack with different shooting distances.}}

\begin{figure}[hptb]
	\centering
	\includegraphics[width=3in]{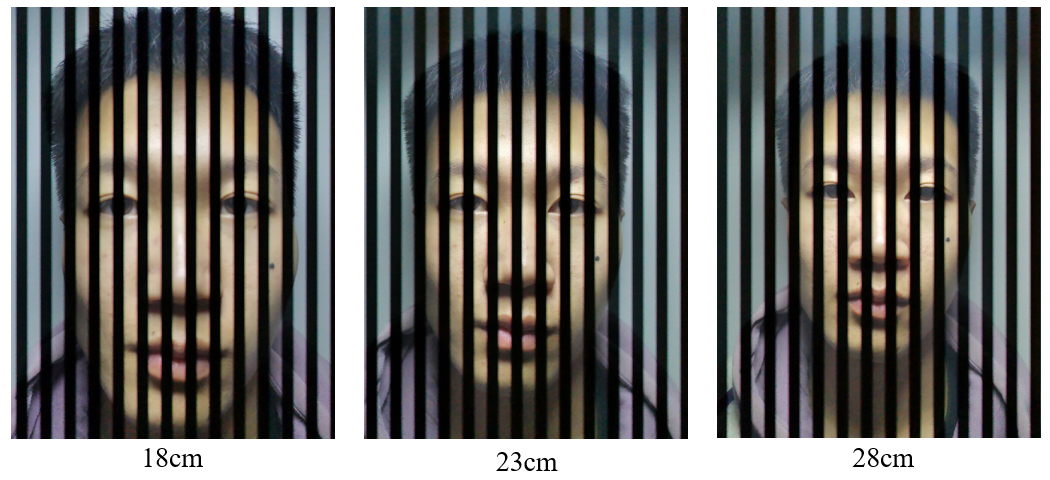}
	\caption{Adversarial examples of DoS attack with different shooting distances.}
	\label{fig_6}
\end{figure}

In real world, attackers are not always able to precisely control the shooting distance between the camera and the face. Since the percentage of facial features in an image changes accordingly with the shooting distance between camera and face, to investigate the effect of difference in the percentage of face in the image on the performance of DoS attack, we collected face images of three volunteers at shooting distances of 18cm (e.g., cell phone face recognition unlock), 23cm (e.g., computer face recognition unlocking), and 28cm (e.g., face recognition access control), 10 images of the faces of three volunteers each, for a total of 180 images. Some of the adversarial examples are shown in Fig. \ref{fig_6}.

The performance of DoS attack with different shooting distances is given in Table \ref{tab4}. We can observe that at different shooting distances, the adversarial examples of DoS attack work stably with an attack success rate higher than 92$\%$, which can threaten face detection models in different application scenarios.

\begin{table}[hbtp]
	\centering
	\caption{The success rate of DoS attack with different shooting distances}
	\label{tab4}
	\renewcommand{\arraystretch}{1.3}
	\begin{tabular}{|l|c|c|c|}
	\hline
	\multirow{2}{*}{Target model} & \multicolumn{3}{c|}{Shooting distance} \\
	\cline{2-4}
     ~	& 18cm        & 23cm       & 28cm       \\
    \hline
	Dlib                          & 97.67\%     & 95.67\%    & 92.67\%    \\
	\hline
	MTCNN                         & 100\%       & 100\%      & 100\%      \\
	\hline
	RetinaFace                    & 100\%       & 100\%      & 100\%     \\
	\hline
	\end{tabular}
\end{table}

(3) {\bf{DoS attack with different tilt angles.}}

\begin{figure}[hptb]
	\centering
	\includegraphics[width=3in]{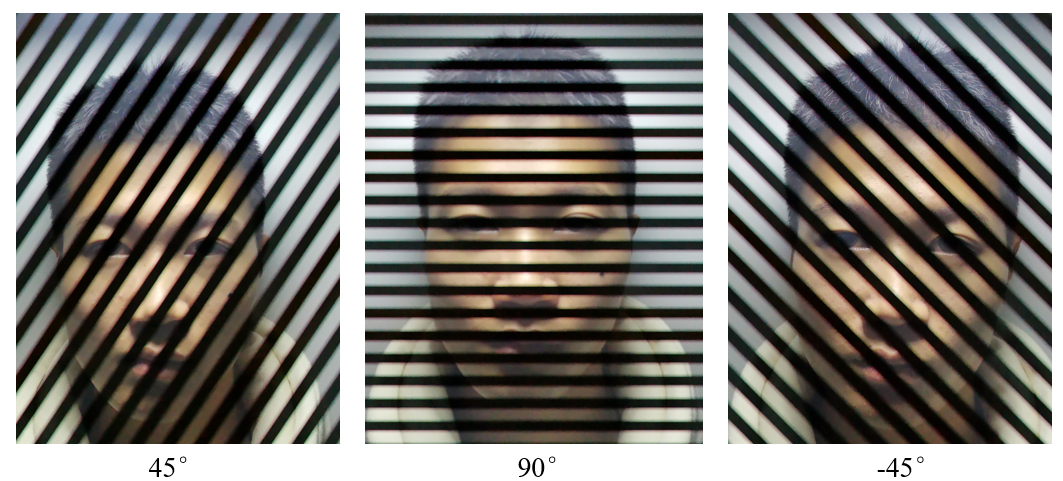}
	\caption{Adversarial examples of DoS attack with different tilt angles.}
	\label{fig_7}
\end{figure}

To test the effect of DoS attacks on the performance of the face recognition system with different adversarial perturbation fringe tilt angles, we collected face images by rotating the camera. We collected three face images of each of the three volunteers at the perturbation fringe tilt angles of 45°, 90°, and -45°, for a total of 54 images, and some of the adversarial examples are shown in Fig. \ref{fig_7}.

Table \ref{tab5} gives the success rate of DoS attack with different tilt angles. With different perturbation fringe tilt angles, we can achieve 100$\%$ success rate of DoS attack for different face detection models,  because the tilted perturbation fringes are more "coherent" to the destruction of facial features, and the dark fringes have a greater range of occlusion for "eyes" and "mouth", or even completely obscured.

\begin{table}[hbtp]
	\centering
	\caption{The success rate of DoS attack with different tilt angles}
	\label{tab5}
	\renewcommand{\arraystretch}{1.3}
	\begin{tabular}{|l|c|c|c|}
		\hline
		\multirow{2}{*}{Target model} & \multicolumn{3}{c|}{Tilt angle} \\
		\cline{2-4}
		~	& 45°        & 90°       & -45°       \\
		\hline
		Dlib                          & 100\%     & 100\%    & 100\%    \\
		\hline
		MTCNN                         & 100\%       & 100\%      & 100\%      \\
		\hline
		RetinaFace                    & 100\%       & 100\%      & 100\%     \\
		\hline
	\end{tabular}
\end{table}

\subsection{Experimental Results of Dodging Attack}
Since fringes with tilted angle are very destructive to the coherent detection of facial features, as shown in the experiment in \uppercase\expandafter{\romannumeral4}-\emph{B}. Therefore, for the dodging attack that interferes with facial features, we only consider the attack with vertical perturbation fringes.

A successful dodging attack against the face feature matching stage means that the face recognition system verifies the face images of two different person with the addition of adversarial bright and dark perturbation fringes as the same person. In our experiments, we input two different face images with bright and dark perturbation fringes into the face verification model, and compare the relationship between the distance of face feature vectors and the verification threshold to determine whether they are verified as the same person. If the number of samples that can be successfully matched before the attack is $m_b$, and the number of adversarial examples that cause face matching errors after dodging attack is $m_a$, then the attack success rate $r_{dodging}$ of the dodging attack is: 
\begin{equation}
	r_{dodging} = ~\frac{m_{a}}{m_{b}}
\end{equation}

In total, we measure the success rate of dodging attack under different pulse repetition periods and different shooting distances.

(1) {\bf{Dodging attack with different pulse repetition periods.}}

We collected 10 images each of the faces of three volunteers under normal LED illumination and modulated LED illumination with pulse repetition periods of 600 $\mu$s and 800 $\mu$s in the environment described in \uppercase\expandafter{\romannumeral4}-\emph{A}, for a total of 90 images, when the camera-to-face shooting distance was fixed at 18 cm.

For the face images under normal illumination, the verification accuracy of all three face verification models is 100$\%$. According to Eq. 15, for each face verification model, we obtain the success rate metrics of the dodging attack in the form of a test-by-test comparison of normal samples and adversarial examples.

\begin{table}[hbtp]
	\centering
	\caption{The success rate of dodging attack with different pulse repetition periods}
	\label{tab6}
	\renewcommand{\arraystretch}{1.3}
	\begin{tabular}{|l|c|c|}
		\hline
		\multirow{2}{*}{Target model} & \multicolumn{2}{c|}{Pulse period} \\
		\cline{2-3}
		~	& 600$\mu$s        & 800 $\mu$s       \\
		\hline
		Dlib                          & 89\%     & 100\%   \\
		\hline
		FaceNet                         & 100\%       & 100\%      \\
		\hline
		ArcFace                    & 69.33\%       & 100\%    \\
		\hline
	\end{tabular}
\end{table}

The success rates of dodging attacks with modulated LED illumination of different pulse repetition periods are given in Table \ref{tab6}. Generally, the modulated LED illumination poses a real threat to all three typical face verification models tested with different pulse repetition periods. However, there is a difference in the effect of finer perturbation fringes on the attack achieved by the different face verification models. And Dlib \cite{ref23} and ArcFace \cite{ref21} have a better ``filtering" effect on the feature interference caused by finer perturbation fringes. When the pulse repetition period of LED modulation is 400$\mu$s, the success rate of dodging attack for all three face verification models reaches 100$\%$.
 
(2) {\bf{Dodging attack with different shooting distances.}}

To investigate the effect of difference in the proportion of face in the image on the performance of dodging attack, we collected the images under normal LED illumination and modulated LED illumination with a pulse repetition period of 800$\mu$s at shooting distances of 18cm, 23cm and 28cm, 10 images each of the faces of three volunteers were collected, for a total of 180 images. 

Table \ref{tab7} gives the success rate of dodging attack with different shooting distances, from which we can see that the adversarial examples of dodging attack work stably with an attack success rate higher than 90$\%$ at different shooting distances, which can effectively threaten face verification models in different application scenarios.

\begin{table}[hbtp]
	\centering
	\caption{The success rate of dodging attack with different shooting distances}
	\label{tab7}
	\renewcommand{\arraystretch}{1.3}
	\begin{tabular}{|l|c|c|c|}
		\hline
		\multirow{2}{*}{Target model} & \multicolumn{3}{c|}{Shooting distance} \\
		\cline{2-4}
		~	& 18cm        & 23cm       & 28cm       \\
		\hline
		Dlib                          & 100\%     & 90.66\%    & 90.32\%    \\
		\hline
		FaceNet                         & 100\%       & 100\%      & 100\%      \\
		\hline
		ArcFace                    & 100\%       & 100\%      & 100\%     \\
		\hline
	\end{tabular}
\end{table}

\subsection{Exploration of potential defense methods for our attacks}
\begin{figure}[hptb]
	\centering
	\includegraphics[width=2.3in]{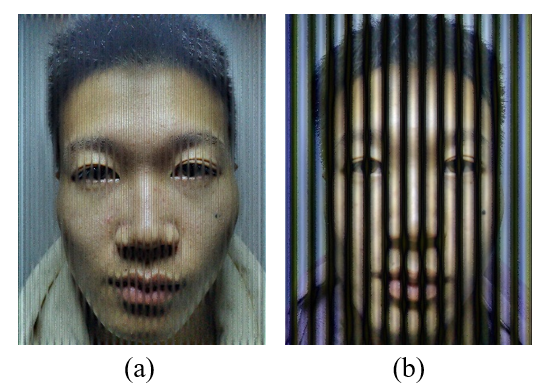}
	\caption{A demonstration of the frequency based defense: (a)600$\mu$s; (b)1800$\mu$s.}
	\label{fig_8}
\end{figure}

For the proposed LIM attacks, we try to present a defense method from frequency domain by adopting the Butterworth filter \cite{ref32} to prevent the signal of specific frequencies (i.e., the luminance information perturbation) from passing through. A demonstration of the result is shown in Fig. \ref{fig_8}.

However, this method is only effective to the case where the perturbation fringes are thin. We conducted experiment on filter-repaired adversarial examples to obtain the defense rate assessment, as shown in Table \ref{tab8}. For DoS attacks with LED pulse periods below 1600 $\mu$s, filter repair enables most of the images to be re-detected faces by the Dlib \cite{ref23} and MTCNN \cite{ref24} model, while most of the filtered images still have a strong attack effect against RetinaFace \cite{ref25}. When the LED pulse period is higher than 1800 $\mu$s, the filtering restoration has a weak effect on the restoration of the adversarial examples, and only the high-frequency signals at the edge of the adversarial fringe perturbation produce a filtering effect, and the filtered images are still unable to detect faces by Dlib \cite{ref23}, MTCNN \cite{ref24}, and RetinaFace \cite{ref25}. For dodging attacks with LED pulse periods of 600 $\mu$s and 800 $\mu$s, the filtered repaired adversarial examples can still pose some threat to Dlib \cite{ref23}, FaceNet \cite{ref21} and ArcFace \cite{ref22}.

\begin{table*}[hbtp]
	\centering
	\caption{The defense rate of filter-repaired adversarial examples}
	\label{tab8}
	\renewcommand{\arraystretch}{1.3}
	\begin{threeparttable}
	\begin{tabular}{|l|c|c|c|c|c|c|c|c|}
		\hline
		\multirow{3}{*}{Target model} & \multicolumn{8}{c|}{Pulse period}                                                         \\
		\cline{2-9}
		& \multicolumn{2}{c|}{Dodging attack} & \multicolumn{6}{c|}{DoS attack}                      \\
		\cline{2-9}
		& 600$\mu$s   & 800$\mu$s  & 1000$\mu$s & 1200$\mu$s & 1400$\mu$s & 1600$\mu$s & 1800$\mu$s & 2000$\mu$s \\
		\hline
		Dlib                          & 33\%             & 48.33\%         & 100\%  & 100\%  & 100\%  & 100\%  & 0      & 0      \\
		\hline
		MTCNN                         & *                & *               & 93.3\% & 100\%  & 96.7\% & 100\%  & 0      & 0      \\
		\hline
		RetinaFace                    & *                & *               & 66.7\% & 26.7\% & 26.7\% & 63.3\% & 0      & 0      \\
		\hline
		FaceNet                       & 52\%             & 15\%            & *      & *      & *      & *      & *      & *      \\
		\hline
		ArcFace                       & 51.33\%          & 12.7\%          & *      & *      & *      & *      & *      & *     \\
		\hline
	\end{tabular}
	\begin{tablenotes}		
		\footnotesize
		\item[*] Not included in the assessment
	\end{tablenotes}
	\end{threeparttable}
\end{table*}

\begin{figure}[hptb]
	\centering
	\includegraphics[width=2.5in]{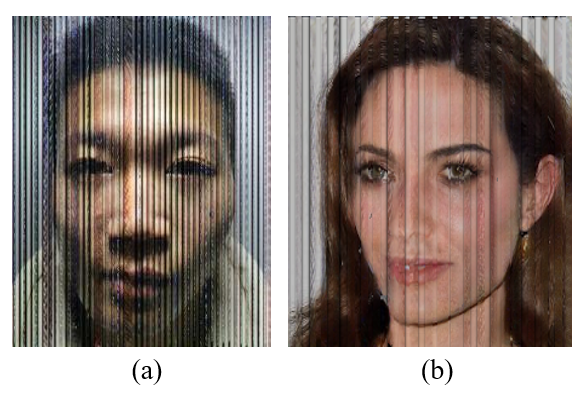}
	\caption{Inpainting adversarial examples: (a)volunteer image; (b)training image.}
	\label{fig_9}
\end{figure}

We also notice that learning-based inpainting method is better for mask-like attack images, so it should also be relevant for our attack method. We use Pluralistic-Inpainting \cite{ref33} method to repair our adversarial examples, but the experimental results show that the repair effect is limited, as shown in Fig. \ref{fig_9}. This learning-based image repair method is not effective for our attack method.

\section{Limitations}
The experiments show that LIM is effective in generating physical adversarial examples for state-of-the-art face recognition systems. However, there are still some limitations.

{\bf{Black-box attack assumptions:}} In this paper, we use a black-box threat model, assuming that attackers only know the output of the face recognition models and have no knowledge of the internal parameters. However, under certain scenarios,  experienced attackers are possibly able to obtain the details of face recognition models. Therefore, LIM can optimize the modulation parameters in more detail by observing the changes of internal parameters of face recognition models, which will help improve the attacking performance.

{\bf{Influence of external lighting:}} Ideally, the modulated LED illumination is the only light source in a face recognition scene. However, the influence of other lighting will inevitably be present in real scenes. Therefore, we implemented the LIM attack in the presence of windows casting external light and additional normal illumination, and found that the adversarial perturbation fringes were still captured normally, and the attack was still effective. As a future work, we will explore the possibility in scenarios with different illumination conditions, and improve the robustness of LIM.

\section{Conclusion}
To conclude, we present a physical adversarial attack method LIM, based on LED illumination modulation to solve the problems of high computational complexity, insufficient concealment and low executability and generalization of attack models in existing physical adversarial attack methods for face recognition systems. The proposed method utilizes the rolling shutter effect of CMOS image sensors in face recognition vision systems to generate imperceptible luminance changes to the human eye through fast intensity modulation of scene LED illumination, thereby perturbing the luminance information into the images acquired by face recognition systems. Based on this, DoS and dodging attack are proposed against face recognition systems. The former can prevent the face recognition system from detecting faces, while the latter can mislead the face recognition system to categorize face images with different identities as the same person. The extensive experimental results on the state-of-art face recognition models demonstrate the effectiveness.

\section*{Acknowledgments}
This work was partially supported by the National Natural Science Foundation of China (No. 62171202), National Science and Technology Major Project Carried on by Shenzhen (CJGJZD20200617103000001), HKU-SCF FinTech Academy, Shenzhen-Hong Kong-Macao Science and Technology Plan Project (Category C Project: SGDX20210823103537030), and Theme-based Research Scheme of RGC, Hong Kong (T35-710/20-R).


\bibliographystyle{IEEEtran}
\bibliography{IEEEexample}

\vfill

\end{document}